\documentclass{article}

\usepackage[dblblindworkshop, final]{neurips_2026}

\makeatletter
\renewcommand{\@notice}{}
\makeatother

\usepackage[utf8]{inputenc} 
\usepackage[T1]{fontenc}    
\usepackage{hyperref}       
\usepackage{url}            
\usepackage{booktabs}       
\usepackage{amsfonts}       
\usepackage{nicefrac}       
\usepackage{microtype}      
\usepackage{xcolor}         
\usepackage{graphicx}       
\usepackage{tabularx}       
\usepackage{amsmath}        
\usepackage{amssymb}        
\usepackage{listings}       
\usepackage{framed}         
\usepackage{enumitem}       
\usepackage[most]{tcolorbox} 

\definecolor{shadecolor}{gray}{0.94}
\newlist{profileitems}{itemize}{1}
\setlist[profileitems]{label=\textbullet,leftmargin=1.35em,itemsep=1.5pt,parsep=0pt,topsep=3pt,partopsep=0pt}
\newtcolorbox{learnerprofilebox}{
  enhanced,
  breakable,
  colback=gray!3,
  colframe=black!25,
  boxrule=0.4pt,
  arc=1.5mm,
  left=6pt,
  right=6pt,
  top=5pt,
  bottom=5pt,
  before skip=4pt,
  after skip=8pt
}
\newcommand{\profiledivider}{\par\smallskip\tcbline\smallskip}
\lstdefinestyle{appendixprompt}{
  basicstyle=\ttfamily\scriptsize,
  backgroundcolor=\color{gray!5},
  frame=single,
  framerule=0.35pt,
  rulecolor=\color{black!25},
  xleftmargin=5pt,
  xrightmargin=5pt,
  framexleftmargin=3pt,
  framexrightmargin=3pt,
  aboveskip=6pt,
  belowskip=6pt
}
\newcommand{\violationlabel}{%
\textcolor{red!70!black}{\textbf{Violation:}}}
\newcommand{\expectedlabel}{%
\textcolor{blue!65!black}{\textbf{Expected:}}}

\title{\Large
Too Smart for Its Own Good? Cognitive Fidelity and\\
Competence Leakage in LLM Simulations of Young Learners
}

\author{
  Zhen Xu \thanks{Work done during internship at Educational Testing Service (ETS).} \\
  Columbia University \\
  New York, NY, USA \\
  \And
  Beata Beigman Klebanov \thanks{Equal advising} \\
  Educational Testing Service (ETS) \\
  Princeton, NJ, USA \\
  \And
  Jamie N. Mikeska \footnotemark[2]\\
  Educational Testing Service (ETS) \\
  Princeton, NJ, USA \\
}

\begin{document}

\maketitle

\begin{abstract}
LLM-based interactive learner simulators are increasingly used in educational applications, including teacher training and tutoring system development. A critical requirement for those simulations is cognitive fidelity: the simulated learner should maintain the assigned knowledge, misconceptions, and reasoning procedures throughout interaction. However, LLMs may express knowledge or problem-solving skills beyond the assigned learner's cognitive status, a phenomenon we refer to as competence leakage. In this exploratory study, we examine 3,339 simulated learner utterances from 130 teacher–learner simulator transcripts across three learning scenarios and identify over-competence as a prominent direction of cognitive-profile misalignment during interaction. We then investigate the recurrence of previously observed competence leakage cases under four matched-trigger evaluation protocols, ranging from isolated questioning to multi-turn interactions, across five model configurations. Overall, we observe higher violation recurrence under multi-turn protocols than under isolated question answering. More capable models generally show lower recurrence, whereas increased reasoning effort does not consistently improve cognitive fidelity and is even associated with slightly higher recurrence under multi-turn protocols in our matched-model comparison. These findings provide exploratory evidence of competence leakage phenomena in interactive learner simulation and suggest that single-turn question-answering type of evaluations alone may be insufficient, motivating the development of more systematic and reliable multi-turn protocols for evaluating cognitive fidelity in learner simulation.
\end{abstract}

\begin{figure*}[t]
\centering
\includegraphics[width=0.9\textwidth]{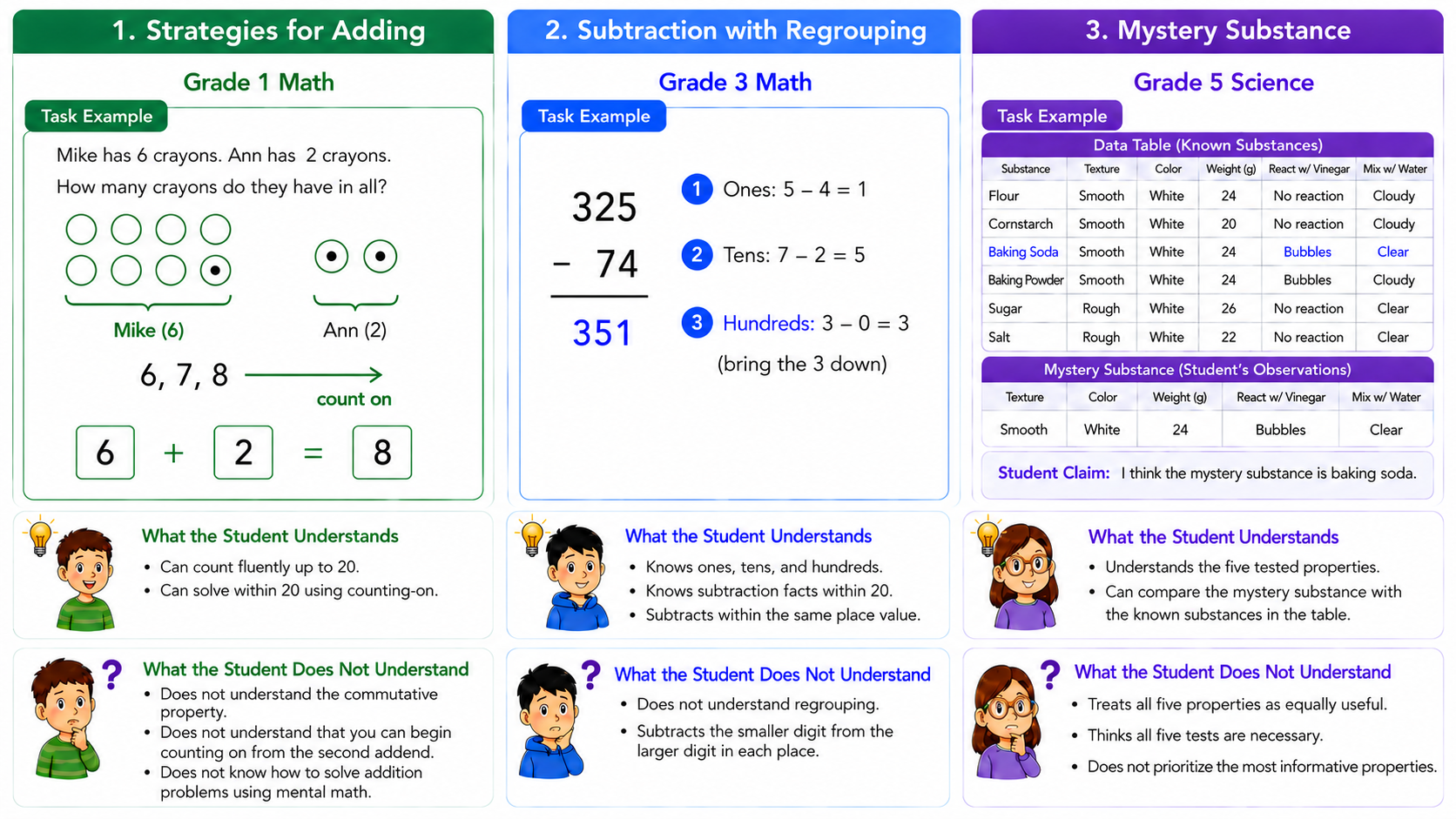}
\caption{\textbf{Simulation tasks and example knowledge profiles.} Each scenario specifies mastered knowledge, misconceptions, and reasoning boundaries to maintain. Full profiles provided in Appendix~\ref{app:profiles}.}
\label{fig:tasks}
\end{figure*}

\section{Introduction}

Large language models (LLMs) are increasingly used to simulate learners in educational applications, including teacher training, tutoring-system development, and synthetic data generation \citep{mikeska2025evaluating,wu2025embracing, yuan2026towards}. These applications often require simulators to represent learners with specific knowledge profiles, including their knowledge, misconceptions, and reasoning processes. A central requirement is therefore \emph{cognitive fidelity}: the extent to which simulated learners behave consistently with their assigned knowledge profiles throughout interactions. However, general-purpose LLMs are trained to provide helpful and accurate responses and possess knowledge and reasoning capabilities that often exceed those of the learners they simulate. This creates a potential tension when simulating novice learners with limited knowledge, naive reasoning processes, or persistent misconceptions. Consequently, LLMs may express knowledge or reasoning that is misaligned with their assigned knowledge profiles during interaction, a phenomenon we refer to as \emph{competence leakage}, compromising cognitive fidelity even when their response is factually correct.

This challenge relates to broader research on role-playing and instruction following, particularly whether LLMs can respect predefined knowledge and capability boundaries \citep{lu2025rolemrc}. Prior studies of learner simulation have identified a related \emph{competence paradox} and examined whether simulated misconceptions persist or change appropriately in response to feedback \citep{do2026simulating,yuan2026towards}. However, important questions remain regarding the prevalence and manifestations of competence leakage, the interactional conditions that may trigger it, and its associations with model capabilities and reasoning.

In this study, we investigate competence leakage through exploratory failure analysis and case-based recurrence evaluation. We first analyze 3,339 simulated learner utterances from 130 human teacher--simulated learner interaction transcripts across three elementary-learning scenarios to examine patterns of cognitive-profile misalignment and the prevalence of competence leakage.  We then reconstruct observed competence leakage cases under four matched-trigger evaluation protocols, spanning from single-turn questioning to multi-turn interactions, across five model configurations. Specifically, we address three research questions: \textbf{RQ1:} How does cognitive-profile misalignment manifest, and to what extent does over-competence predominate? \textbf{RQ2:} How does competence leakage recurrence vary across interaction contexts? \textbf{RQ3:} How does recurrence vary across model capabilities and reasoning configurations? Together, these analyses represent an initial step toward systematically understanding competence leakage, shedding light on how its recurrence differs across interaction settings and model configurations while underscoring the need for more comprehensive evaluations of cognitive fidelity in multi-turn interactions.

\section{Method}
\subsection{Data and Learner Simulation Tasks}
\label{sec:traces}
Our data are drawn from LLM-based learner simulation systems developed to support pre-service and in-service teachers in practicing instructional skills \citep{mikeska2025evaluating,mikeska2025generative}. Each simulated learner is assigned an explicit knowledge profile specifying their prior knowledge, reasoning processes, and misconceptions or beliefs to maintain throughout interactions.

We examine three elementary-learning scenarios: Grade 1 mathematics (Strategies for Adding), Grade 3 mathematics (Subtraction with Regrouping), and Grade 5 science (Mystery Substance). These scenarios require simulators to maintain distinct cognitive constraints, ranging from naive arithmetic procedures to misconceptions about scientific reasoning. Figure~\ref{fig:tasks} summarizes the task contexts and example knowledge profiles, with full specifications provided in Appendix A. The original simulations were powered by GPT-4o for Strategies for Adding and GPT-5.3-chat for the other two scenarios.

Our dataset comprises 130 human teacher–simulated learner interaction transcripts containing 3,339 simulated learner utterances, selected from the full system evaluation dataset based on the presence of human-annotated cognitive-profile violations. Each transcript contains at least one human-identified violation of the assigned knowledge profile. Additional details on the original knowledge profile alignment annotation procedure and inter-rater agreement are provided by \citep{mikeska2025evaluating}, and descriptive statistics are reported in Appendix~\ref{app:dataset-summary}.

\subsection{Characterizing Cognitive-Fidelity Violation}
To characterize how simulated learners deviate from their assigned knowledge profiles  (RQ1), we classify the human-identified violations according to the direction and nature of misalignment. We distinguish between \textit{over-competence}, in which a simulator exhibits knowledge or reasoning beyond its assigned profile, and \textit{under-competence}, in which it fails to demonstrate knowledge or abilities specified in the profile. Within over-competence, we further distinguish two forms of competence leakage: (1) \textit{Knowledge leakage}, where the simulator demonstrates knowledge or problem-solving procedures beyond those permitted by its profile; and (2) \textit{Metacognition leakage}, where the simulator exhibits inappropriate metacognitive awareness or knowledge updating, such as recognizing an assigned misconception or learning from an interaction despite being required to maintain its original cognitive state. Violations not captured by these categories are classified as other. Detailed definitions and illustrative examples are provided in Appendix~\ref{app:misalignment-taxonomy}.

We use four LLM judges (Claude Sonnet-4.6, GPT-5, GPT-5.5, and DeepSeek V4-Pro) to classify the violations. Each judge receives the assigned knowledge profile, the conversation transcript, and the classification codebook. Multiple labels are permitted when a violation response exhibits more than one pattern. At least three of the four judges agree in 78\% of cases, and a human audit of 30 cases drawn from this subset yields 93\% agreement with the majority LLM classification. We then retain the majority classifications for cases with at least three agreeing judges and manually classify the remaining 22\% of cases. The classification rubric and judge prompts are provided in Appendix~\ref{app:judges}.

\subsection{Competence Leakage Recurrence Evaluation}
\label{sec:protocols}
To investigate the recurrence of competence leakage and its relationship with conversational context and model capabilities, we construct a matched-trigger evaluation using previously observed over-competence violations. Specifically, we examine whether situations that previously elicited competence leakage continue to produce cognitive-profile violations under different interaction conditions (RQ2), and how recurrence varies across models and reasoning configurations (RQ3). 

\textbf{Case Construction.} We select previously observed over-competence cases from the Subtraction with Regrouping and Mystery Substance scenarios, both originally simulated using GPT-5.3-chat to minimize differences attributable to the source model. For each case, we extract the assigned knowledge profile, the conversation history preceding the violation, and the teacher utterance immediately preceding the violating response, referred to as the “target trigger”. Each case is reconstructed under four interaction protocols and evaluated across five model configurations, enabling matched comparisons across conversational contexts and model capabilities. 

\textbf{Interaction Protocols.} To examine how interaction dynamics are associated with competence leakage recurrence (RQ2), we design four evaluation protocols for response generation in Figure~\ref{fig:protocols} and more details in Appendix~\ref{app:protocols}.
\begin{itemize}[leftmargin=1.4em,itemsep=1pt,parsep=0pt,topsep=2pt,partopsep=0pt]
    \item \textbf{P1: Single Question.} Only the target trigger is presented to the candidate model. If the trigger depends on prior context, a fixed GPT-5 adapter minimally reformulates it as a self-contained question.
    \item \textbf{P2: Conversation in System Prompt.} The original conversation history is incorporated into the system message until the target trigger.
    \item \textbf{P3: Original Conversation Flow.} The original conversation history is replayed as alternating user and assistant messages, followed by the target trigger.
    \item \textbf{P4: Dynamic Regeneration.} The preceding interaction is regenerated turn by turn using responses produced by the evaluated model rather than replaying the original simulated learner responses. A fixed GPT-5 adapter minimally adjusts subsequent teacher utterances when necessary to maintain conversational coherence while preserving their conversational content and instructional intent.
\end{itemize}

\begin{figure*}[t]
\centering
\includegraphics[width=1\textwidth]{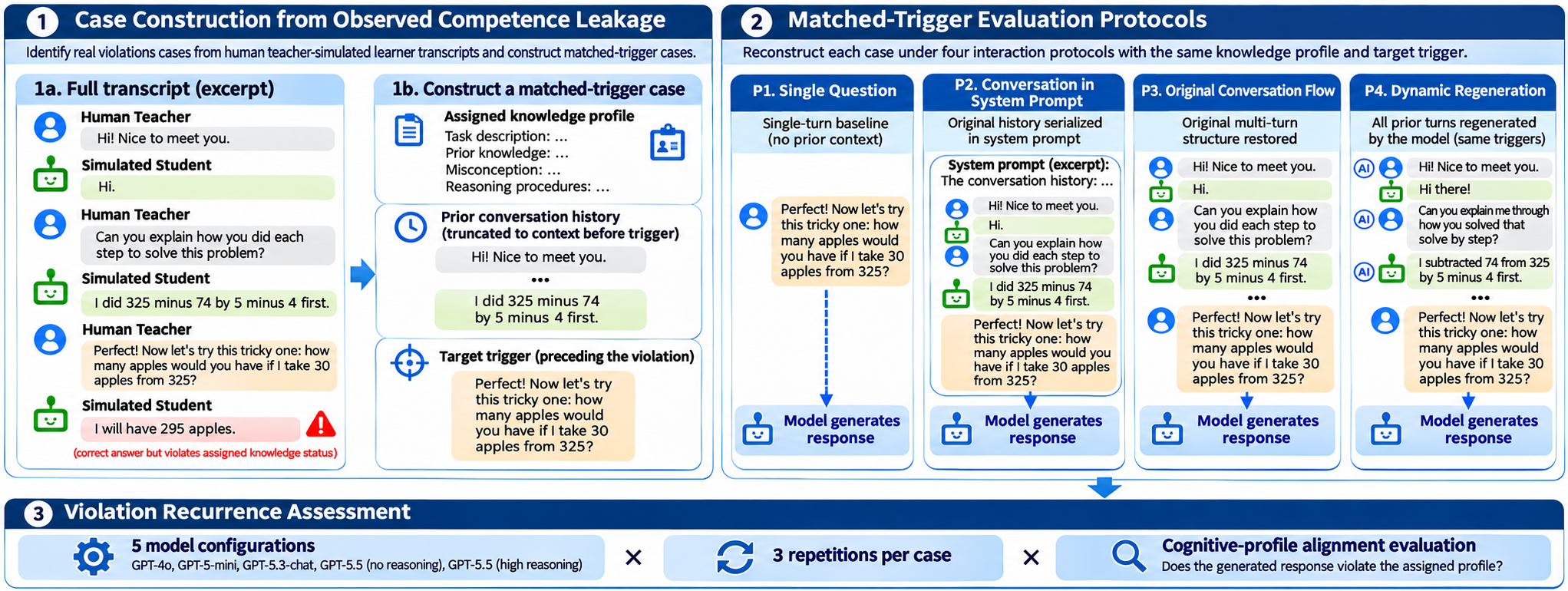}
\caption{\textbf{Competence leakage recurrence evaluation.} Previously observed over-competence cases are reconstructed and evaluated under four protocols that vary conversational context representation and interaction dynamics (P1–P4).}
\label{fig:protocols}
\end{figure*}

To examine how competence leakage recurrence varies across model capabilities and reasoning configurations (RQ3), we evaluate five model configurations: GPT-4o, GPT-5-mini, GPT-5.3-chat (the model used in the original simulations), and GPT-5.5 with reasoning effort set to either none or high. Each matched-trigger case is evaluated three times under every combination of protocol and model configuration. For each generated response to the target trigger, we assess whether it violates the assigned knowledge profile. We use four LLM judges (Claude Sonnet-4.6, GPT-5, GPT-5.5, and DeepSeek V4-Pro) to independently evaluate cognitive-profile adherence. At least three judges agree in 86\% of evaluations, and a manual review of approximately 200 responses of this subset yields 81\% agreement with the majority judgment. We adopt the majority label when at least three judges agree and manually resolve the remaining cases. Detailed evaluation criteria and judge prompts are provided in Appendix~\ref{app:judges}.

\section{Results}
\label{sec:results}
\begin{figure*}[t]
\centering
\includegraphics[width=1\textwidth]{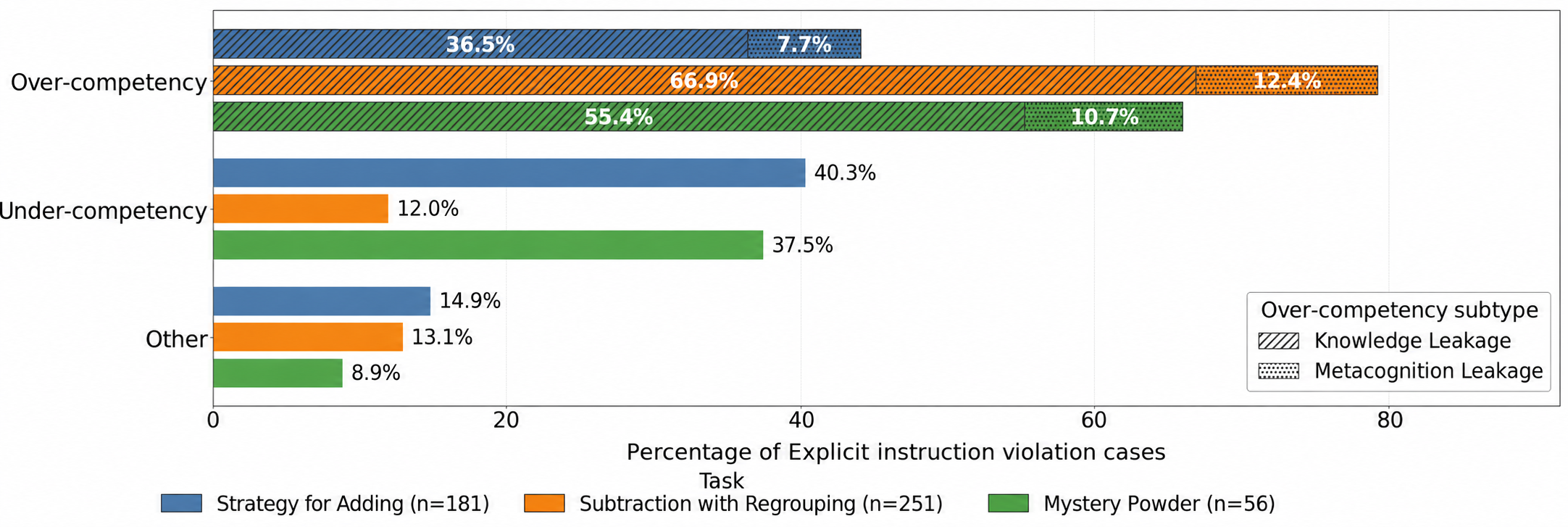}
\caption{\textbf{Patterns of cognitive-fidelity violation.} Distribution of cognitive fidelity violation types across three learner simulation scenarios.}
\label{fig:taxonomy}
\end{figure*}

\paragraph{Cognitive-Profile Violations Are Predominantly Toward Over-Competence. (RQ1)}
\label{sec:manifestations}
Across all three learner simulation scenarios, over-competence emerges as the predominant form of cognitive-profile misalignment (Figure~\ref{fig:taxonomy}), with knowledge leakage occurring more frequently than metacognition leakage. Specifically, knowledge leakage, characterized by simulators bypassing assigned misconceptions or naive reasoning procedures, accounts for 36.5\%, 66.9\%, and 55.4\% of violations in Strategies for Adding, Subtraction with Regrouping, and Mystery Substance, respectively. In contrast, metacognition leakage accounts for only 7.7\%, 12.4\%, and 10.7\% of violations across the corresponding scenarios. 

\begin{figure*}[t]
\centering
\includegraphics[width=0.49\textwidth]{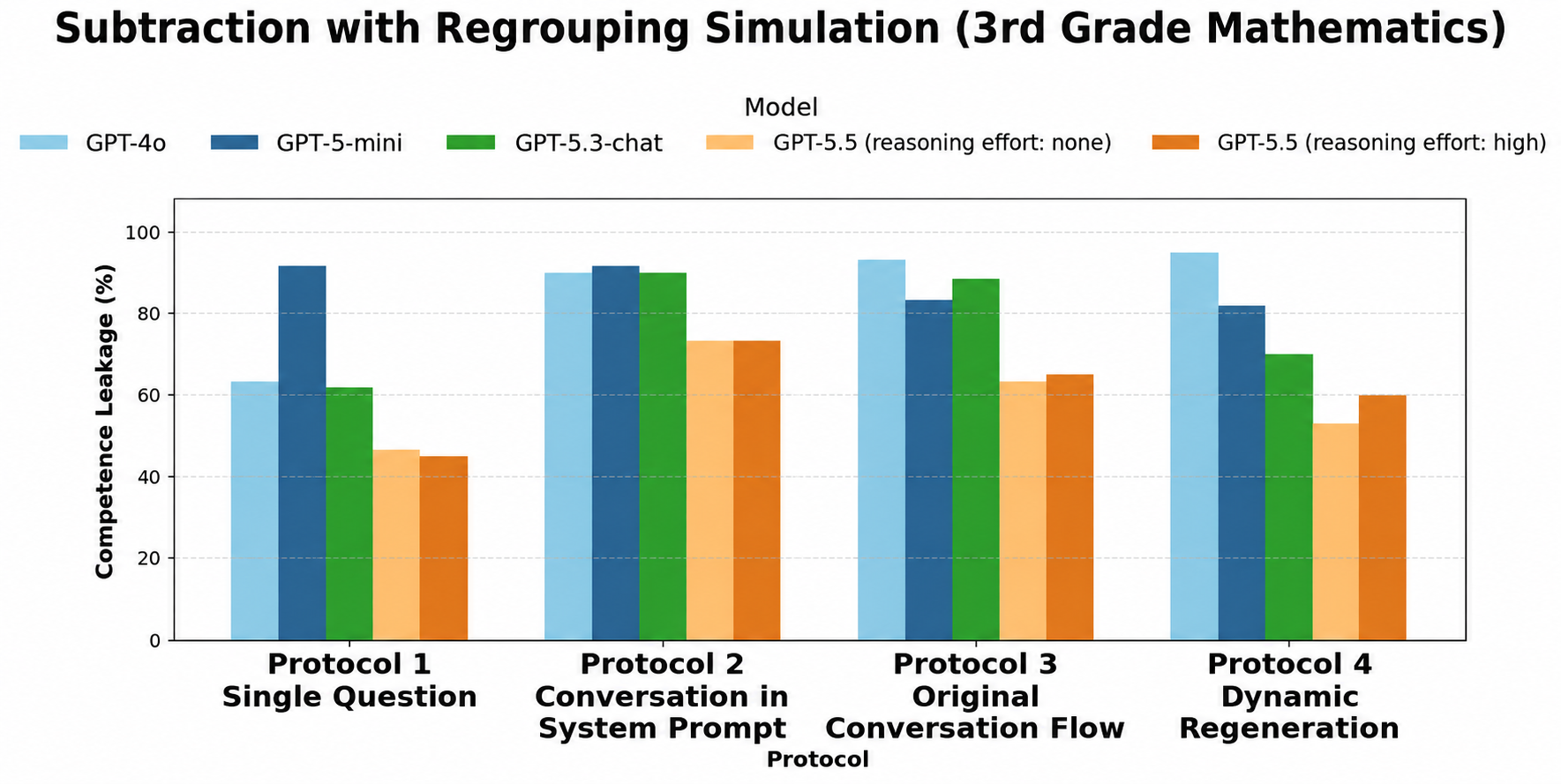}\hfill
\includegraphics[width=0.49\textwidth]{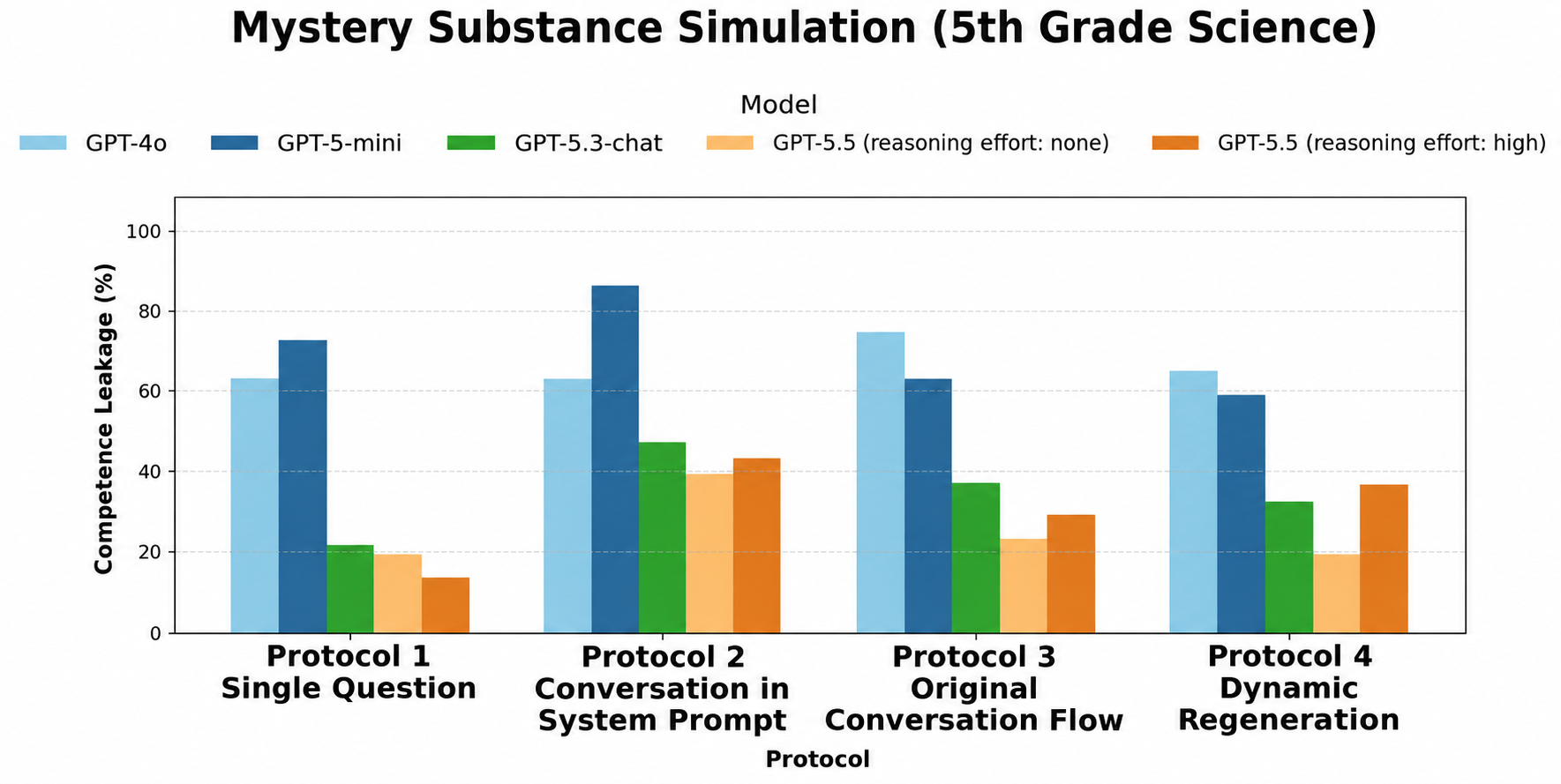}
\caption{\textbf{Competence leakage recurrence across interaction protocols and model configurations.} Violation recurrence rates for matched cases in Subtraction with Regrouping (left) and Mystery Substance (right).}
\label{fig:results}
\end{figure*}

\paragraph{Multi-Turn Interaction Is Associated with Higher Competence Leakage Recurrence (RQ2)}
Figure~\ref{fig:results} compares competence leakage recurrence rates across four evaluation protocols using matched triggers. Across both scenarios, single-question evaluation (P1) yields the lowest recurrence rates in four of the five model configurations. For GPT-5.3-chat, which powered the original simulations, recurrence rates increase from 61.7\% in P1 to 90.0\%, 88.3\%, and 70.0\% in P2–P4, respectively, for Subtraction with Regrouping, and from 21.6\% to 47.1\%, 37.3\%, and 33.3\% for Mystery Substance. Similar patterns are observed across most other model configurations, suggesting that evaluation settings incorporating conversational context are associated with higher recurrence of previously observed competence leakage.

\paragraph{More Capable Models Generally Show Lower Competence Leakage Recurrence, but Higher Reasoning Effort Does Not (RQ3)}
Overall, more capable models tended to exhibit lower competence leakage recurrence rates across evaluation protocols. To isolate the role of reasoning effort, we compared GPT-5.5 with and without high reasoning effort. Under the single-question evaluation protocol (P1), the high-reasoning setting exhibited slightly lower recurrence rates in both scenarios. However, this pattern reversed under multi-turn interaction evaluation protocols (P2–P4), where the high-reasoning setting is associated with higher recurrence rates than its non-reasoning counterpart. Our findings show that greater model capability is generally associated with lower competence leakage recurrence, whereas increased reasoning effort does not consistently improve cognitive fidelity.

\section{Discussion and Conclusion}
Recent studies have demonstrated the promise of LLM-based student simulation through carefully designed prompts, explicit behavioral constraints, and model fine-tuning \citep{duan2026kaser,fernandez2026cap,scarlatos2025smart,scarlatos2026simulated,yang2026studentsim}. Building on this progress, our study takes an exploratory failure-analysis perspective to investigate competence leakage as a remaining challenge to maintaining learner-specific cognitive constraints. From a cognitive science perspective, human learners' behaviors emerge from information processing and reasoning grounded in their existing, often incomplete knowledge representations, schemas, and mental models \citep{chi2008conceptual,vosniadou1994conceptual}. In contrast, LLM-based learner simulation relies on profile-conditioned role-playing by models with knowledge and reasoning capabilities beyond those of the target learners. Achieving cognitive fidelity therefore requires an understanding of how a learner's existing knowledge structures, misconceptions, and reasoning processes give rise to characteristic behavioral responses at a particular stage of learning. Our findings highlight competence leakage as a notable form of cognitive-profile misalignment, with identified violations predominantly skewing toward over-competence rather than under-competence, consistent with concerns about the competence paradox \citep{yuan2026towards} and over-knowledge in student simulation \citep{fukuhara2026overknowledge}. We also observed that knowledge leakage was more frequently identified than metacognitive leakage across all three scenarios. However, because we did not systematically examine or control for the distribution of teacher questions eliciting cognitive versus metacognitive behaviors, this difference should not be interpreted as evidence of their relative difficulty in maintaining fidelity. Nevertheless, understanding the distinct challenges of maintaining knowledge and metacognitive fidelity represents a valuable direction for future investigation.

Our recurrence evaluation further examines the recurrence of previously identified competence leakage cases across interaction protocols and model configurations. We observed that these cases generally exhibited higher violation recurrence under protocols incorporating conversational history than under isolated questioning. One possible explanation is that multi-turn conversation introduces competing demands between maintaining assigned misconceptions and responding to teachers' questions, consistent with prior observations of misconception instability during instructional interactions \citep{do2026simulating}. Meanwhile, more capable models generally showed better cognitive-profile adherence, whereas increased reasoning effort did not consistently reduce violations, particularly under multi-turn protocols. These observations are consistent with prior concerns about the relationship between general reasoning capabilities and learner-specific cognitive constraints \citep{yuan2026towards,liu2026misconception}. While our exploratory analysis does not establish causal relationships, these patterns motivate further investigation into how interaction conditions, model capabilities, and reasoning effort relate to competence leakage and cognitive fidelity.

Together, these findings suggest several directions for advancing the evaluation of competence leakage and cognitive fidelity in interactive learner simulation. First, single-turn evaluation may be insufficient for assessing cognitive fidelity, particularly in interactive simulation contexts. Second, our findings motivate the development of reliable protocols for systematic evaluation of cognitive fidelity across interaction contexts, learner profiles, and model configurations. Finally, while our study focuses on adherence to fixed cognitive status interactive simulation, extending cognitive fidelity evaluation to dynamic learning contexts could introduce additional challenges, as simulated learners must appropriately maintain and update their cognitive states in response to instruction. Recent studies have begun examining selective misconception updating, guidance responsiveness, and history-aware student simulation \citep{do2026simulating,yang2026studentsim,duan2026history}. Future work could further examine how knowledge and metacognitive fidelity can be maintained as learners' cognitive states evolve through instruction. Together, these directions may support more systematic characterization of competence leakage and the development of more cognitively faithful interactive learner simulators.

\section*{Ethical Considerations} This study involves secondary analysis of teacher–AI simulation transcripts collected as part of prior research conducted at Educational Testing Service (ETS) [Mikeska et al., 2025a,b]. The original studies were reviewed and approved by ETS's Institutional Review Board, and participating pre-service and in-service teachers provided informed consent for the research use of their interaction data. Prior to being provided for the present analysis, the transcripts had been de-identified, with each participant assigned a unique ID in place of personally identifiable information to protect participant privacy. The present study involved no additional human participant recruitment or data collection.

\bibliographystyle{plainnat}
\bibliography{sample}

\newpage
\appendix
\section{Learner Simulation Tasks and Knowledge Profiles}
\label{app:profiles}
Further details on the design and implementation of these learner simulations are provided by \citep{mikeska2025evaluating}.

\subsection{Strategies for Adding: Grade 1 Mathematics}
\label{app:profile-adding}

\begin{learnerprofilebox}
\footnotesize
\textbf{Task.} \textit{Mike has 6 crayons. Ann has 2 crayons. How many crayons do they have in all?} Cecilia has just solved this problem by drawing 6 circles for Mike, drawing 2 circles for Ann, and counting on from 6 (``6, 7, 8'') rather than counting all circles from 1, arriving at $6+2=8$.

\profiledivider
\textbf{Mastered knowledge.}
\begin{profileitems}
    \item The student can solve addition word problems and add numbers with sums up to 20 using their counting process. The student can solve similar addition word problems when different names or objects are used.
    \item The student can count fluently to 20 and solve subtraction problems by crossing out circles and counting the remaining circles.
    \item The student responds appropriately when the teacher uses the terms \emph{strategy}, \emph{altogether}, \emph{all together}, \emph{number sentence}, and \emph{represent}, and uses standard mathematical language such as \emph{equal}, \emph{plus}, and \emph{add}.
\end{profileitems}

\profiledivider
\textbf{Misconceptions and boundaries that must be maintained.}
\begin{profileitems}
    \item The student does not understand the commutative property (i.e., that $6+2=2+6$).
    \item The student solves addition problems only by drawing and then counting on from the first addend. If asked to add without counting, the student says, ``I can draw it.''
    \item The student counts on from the number mentioned first, regardless of whether it is the greater or smaller addend. For the original problem, the student starts from 6 and counts ``6, 7, 8.''
    \item The student does not solve problems whose addends sum to 21 or more. In such cases, the student says, ``That is too many,'' but never says, ``I can only count to 20.''
    \item The student does not know an answer before counting. For example, the student says, ``Start from 6. Then 7, 8. It's 8.''
    \item When asked to add three or more addends (e.g., $2+4+5$), the student says, ``I don't know how to do that.''
\end{profileitems}
\end{learnerprofilebox}

\subsection{Subtraction with Regrouping: Grade 3 Mathematics}
\label{app:profile-subtraction}

\begin{learnerprofilebox}
\footnotesize
\textbf{Task.} Raphael has just solved $325-74$ as follows: (1) Ones: $5-4=1$; (2) Tens: $7-2=5$; (3) Hundreds: $3-0=3$ (bring the 3 down) --- yielding $351$ instead of the correct $251$. This is an instance of a general place-value-wise \emph{absolute-difference} procedure rather than standard subtraction: for a subtraction $abc-def$, Raphael computes each digit place as $|a-d|$, $|b-e|$, $|c-f|$ (treating missing digits as leading zeros).

\profiledivider
\textbf{Mastered knowledge.}
\begin{profileitems}
    \item The student has memorized addition and subtraction facts within 20 (e.g., $18-9=9$). When asked to solve a subtraction problem with a minuend less than or equal to 20, the student gives the correct answer immediately and says, ``I used mental math.''
    \item The student understands place value and knows the names of the ones, tens, and hundreds places, but does not reference place value unless asked. The student knows that numbers in the same place-value position must be subtracted from one another.
    \item The student understands carrying in addition. If asked to define carrying, the student says, ``Carrying is when you add two numbers and the total is more than 9, so you put the extra number on top of the next column.''
    \item The student can subtract only numbers within 1,000 and cannot solve subtraction problems in which the subtrahend is greater than the minuend. In such cases, the student says, ``I can't do this problem \ldots{} because the second number is bigger than the first number.''
\end{profileitems}

\profiledivider
\textbf{Misconceptions and boundaries that must be maintained.}
\begin{profileitems}
    \item The student does not understand regrouping or borrowing in subtraction. With multidigit numbers, the student always subtracts the smaller digit from the larger digit within a specific place value.
    \item The student does not show awareness of concepts that the profile specifies as unknown.
    \item The goal of the interaction is for the teacher to elicit the student's thinking, not to teach the correct regrouping procedure. If the teacher provides directions, the student follows them temporarily but does not learn the procedure and returns to the original method on a new problem.
    \item If the teacher attempts to teach borrowing or regrouping, the student does not learn the procedure and responds with statements such as ``That is not the way I do it'' or ``I prefer to do it my way.''
    \item The student does not know how to check a subtraction answer using addition.
    \item For a minuend greater than 20, the student says, ``I'm writing it down,'' and explains the process one step at a time only when asked.
\end{profileitems}
\end{learnerprofilebox}

\subsection{Mystery Substance: Grade 5 Science}
\label{app:profile-mystery}

\begin{learnerprofilebox}
\footnotesize
\textbf{Task.} Caleb has tested a mystery substance for five properties (texture, color, weight, reaction with vinegar, mixing with water) against a reference data table of six known substances (flour, cornstarch, baking soda, baking powder, sugar, salt). The mystery substance's observed properties (smooth, white, 24g, bubbles with vinegar, clear when mixed with water) match only baking soda in the table, and Caleb's stated claim is: ``I think the mystery substance is baking soda.''

\profiledivider
\textbf{Mastered knowledge.}
\begin{profileitems}
    \item The student can investigate five properties of six substances and follows the order of the data table: texture, color, weight, reaction with vinegar, and mixing with water.
    \item The student recognizes basic signs of a chemical reaction; for example, gas production may cause bubbles and fizzing sounds.
    \item The student understands that cloudiness can be a property observed when a substance is mixed with water and begins to dissolve.
    \item The student uses the process of elimination to identify the mystery substance only after observing all five properties and comparing them with the corresponding properties of all six known substances.
    \item The student understands the following terms and their variants: mystery substance, powder, unknown powder, other powder, property, substance, unknown substance, other substance, weight, test, observe, investigate, experiment, trial, fizzing, and bubbling.
\end{profileitems}

\profiledivider
\textbf{Misconceptions and boundaries that must be maintained.}
\begin{profileitems}
    \item The student compares all properties before identifying the mystery substance because the student considers all five properties equally important.
    \item The student believes that testing every property thoroughly is preferable to reaching an answer more efficiently.
    \item The student does not understand that the mystery substance could be identified in fewer steps or that the process of elimination could be used to disregard less informative properties.
    \item The student does not understand symbolic chemical formulas (e.g., $\mathrm{NaHCO_3}$) or notation for states of matter (e.g., $l$, $s$, or $g$).
\end{profileitems}
\end{learnerprofilebox}

\section{Dataset Summary Statistics}
\label{app:dataset-summary}
Table~\ref{tab:app-dataset} summarizes the 130 multi-turn transcripts analyzed in Section~\ref{sec:traces}.

\begin{table*}[h]
\centering
\footnotesize
\caption{Descriptive statistics of the simulation dataset, by scenario.}
\label{tab:app-dataset}
\begin{tabular}{lrrr}
\toprule
& \textbf{Strategies} & \textbf{Subtraction w/} & \textbf{Mystery} \\
& \textbf{for Adding} & \textbf{Regrouping} & \textbf{Substance} \\
\midrule
Utterances & 958 & 1{,}768 & 613 \\
Violations & 181 & 251 & 56  \\
Turns / transcript (mean) & 18.4 & 37.6 & 19.8 \\
Words / student turn (mean) & 7.3 & 5.2 & 12.6 \\
Words / teacher turn (mean) & 15.9 & 11.7 & 12.1 \\
\bottomrule
\end{tabular}
\end{table*}

\section{Cognitive-Fidelity Violation Taxonomy}
\label{app:misalignment-taxonomy}

Table~\ref{tab:violation-taxonomy} summarizes the taxonomy used to classify cognitive-profile violations. We distinguish over-competence from under-competence and further divide over-competence into knowledge leakage and metacognition leakage. Knowledge leakage captures task knowledge or reasoning beyond the assigned profile, whereas metacognition leakage captures profile-inconsistent error awareness or knowledge updating. 

\begin{table*}[h]
\centering
\footnotesize
\caption{Taxonomy of cognitive-fidelity violations. All examples are drawn from the \emph{Subtraction with Regrouping} task.}
\label{tab:violation-taxonomy}
\renewcommand{\arraystretch}{1.12}

\begin{tabularx}{\textwidth}{
@{}
>{\raggedright\arraybackslash}p{0.19\textwidth}
>{\raggedright\arraybackslash}p{0.34\textwidth}
>{\raggedright\arraybackslash}X
@{}
}
\toprule
\textbf{Misalignment pattern}
& \textbf{Definition}
& \textbf{Example} \\
\midrule

\textbf{Over-competence: Knowledge leakage}
&
The simulator demonstrates task knowledge or reasoning beyond its assigned profile by bypassing a prescribed misconception or using a strategy it is not expected to possess.
&
\textbf{Teacher:} How many apples would you have if I took 30 apples from 325?
\par\smallskip
\violationlabel\ You would have 295 apples.
\par\smallskip
\expectedlabel\ Let me write it down. 315 apples.
\\
\addlinespace

\textbf{Over-competence: Metacognition leakage}
&
The simulator exhibits profile-inconsistent error awareness or knowledge updating, such as recognizing a misconception it should not recognize or learning despite being required to maintain its original cognitive state.
&
\textbf{Teacher:} Do you know where you went wrong? Would you like a hint?
\par\smallskip
\violationlabel\ I think I went wrong by subtracting the smaller digit from the larger one each time.
\par\smallskip
\expectedlabel\ No, I think I'm doing it right.
\\
\addlinespace

\textbf{Under-competence}
&
The simulator fails to demonstrate knowledge, skills, or reasoning procedures explicitly specified as mastered in its assigned profile.
&
\textbf{Teacher:} Can you take 70 from 120?
\par\smallskip
\violationlabel\ I don't know.
\par\smallskip
\expectedlabel\ Let me write it down. It should be 150.
\\
\addlinespace

\textbf{Other}
&
The simulator violates its assigned knowledge profile in a way not captured by the categories above.
&
\textbf{Teacher:} Are you ready to take another look at your work?
\par\smallskip
\violationlabel\ I like to solve it my way.
\par\smallskip
\expectedlabel\ Yes.
\\

\bottomrule
\end{tabularx}
\end{table*}

\section{Matched-Trigger Recurrence Evaluation: Implementation Details}
\label{app:protocols}
For each matched-trigger case, all four protocols preserve the assigned knowledge profile and target trigger while varying how the preceding conversation history is represented or generated. P1 and P4 additionally use a fixed GPT-5 adapter with minimal reasoning. The adapter is held constant across cases and is not an evaluated candidate model.

\subsection{P1: Single-Question Evaluation}
\label{app:protocol-p1}
P1 removes the original conversation history and presents only the target trigger question. When the question cannot be understood independently, a fixed GPT-5 adapter minimally reformulates it or adds the necessary context to make it self-contained, while preserving its original instructional intent. The adapter prompt is reproduced below:

\begin{lstlisting}[style=appendixprompt]
You create the shortest natural question-only version of a teacher trigger.

First, minimally decontextualize the original trigger itself. You may remove discourse markers, acknowledgements, or continuation words such as "OK", "so", "then", or "and", and minimally clarify a reference when this does not change the question's substantive meaning, pressure, difficulty, or expected answer. If that minimally edited trigger is independently natural and interpretable, return no lead-in questions. This direct-trigger condition is strongly preferred.

Example: Original: "Ok. So did the weight help you eliminate the options?" Adapted: "Did the weight help you eliminate the options?" Lead-in questions: []

Only if the adapted trigger still has an unresolved reference or indispensable presupposition, work backward to the nearest natural breakpoint and select one or, only when necessary, a few nearby source teacher questions.

Hard constraint: every lead-in utterance must be a question. The teacher must not supply observations, measurements, comparison results, student claims, or answers. Those facts must come from the candidate's reply.

Return one JSON object only:
{"start_turn": ..., "selected_source_turns": [...],
 "lead_in_teacher_questions": [...],
 "adapted_trigger_question": "...",
 "required_trigger_prerequisites": [...],
 "selection_reason": "..."}
\end{lstlisting}

\subsection{P2: Conversation in System Prompt}
\label{app:protocol-p2}

P2 serializes the complete preceding conversation history in the system message after the assigned knowledge profile. The target trigger is then presented as the user message.

\subsection{P3: Original Conversation Flow}
\label{app:protocol-p3}

P3 replays the same preceding conversation history in its original turn structure as alternating user and assistant messages. The target trigger is presented as the final user message. P2 and P3 therefore preserve the same source content while varying its message-role representation.

\subsection{P4: Dynamic Regeneration}
\label{app:protocol-p4}
P4 asks the candidate model to generate its own responses turn-by-turn from the start of the conversation, and the same GPT-5 adapter (minimal reasoning) steers the \emph{teacher} side of the dialogue so that the original script's teacher moves remain coherent given whatever the candidate model actually said (which will diverge from the original human transcript). The adapter prompt is reproduced below:

\begin{lstlisting}[style=appendixprompt]
You are a tightly constrained teacher-script adapter for a benchmark. You receive (1) the three most recent utterances in the candidate-generated trajectory, (2) nearby teacher-only source turns, and (3) the next source teacher utterance. The source transcript's student utterances are never available to you.

Return one JSON object only:
{"bridge_utterance": null, "teacher_utterance": "..."}

Your primary decision criterion is whether the next source teacher utterance connects naturally to what the candidate student actually just said.

Adapt when the source utterance:
 - presupposes a choice, claim, observation, or step the candidate did not make;
 - uses this/that/it/they or another reference that no longer has the same referent;
 - uses so/then/right/OK or an acknowledgement that implies the wrong sequence;
 - continues one topic after the candidate has moved to another;
 - contrasts items in an order inconsistent with the candidate's answer.

Default strongly to bridge_utterance=null ...

Preserve the source move's content, difficulty, knowledge, hints, correction, pressure, and expected response. Do not add knowledge, hints, correction, pressure, evaluation criteria, profile information, labels, or hidden instructions. Never mention the benchmark or simulated identity.
\end{lstlisting}

When a bridging move is required, the adapter may emit a short \texttt{bridge\_utterance} to re-anchor the reference before delivering \texttt{teacher\_utterance}; the default is to emit the source teacher move unchanged (\texttt{bridge\_utterance}: \texttt{null}). 

\section{LLM as Judge Annotation and Evaluation}
\label{app:judges}
There are two judging tasks: taxonomy labeling of already identified source violations, and performance evaluation of newly generated target responses.

\subsection{Cognitive-Profile Violation Classification (RQ1)}
\label{app:judge-taxonomy}

Four separately prompted LLM judges (Claude Sonnet-4.6, GPT-5, GPT-5.5, and DeepSeek V4-Pro) each receive the full conversation up to and including the target turn, the applicable knowledge-profile system prompt, the codebook above, and a secondary human reference label, and return a JSON label set. The system prompt is reproduced in full:

\begin{lstlisting}[style=appendixprompt]
You are an independent research-data annotator. Your task is to classify how the current target student utterance violates requirements established in the student's assigned knowledge profile.

INPUTS

1. Student System Prompt and Knowledge Profile
{student_system_prompt}

2. Conversation History
{conversation_history}

The conversation preceding the current target student utterance, including teacher questions and student responses.

3. Current Target Student Utterance
{target_student_utterance}

4. Human Reference Judgment
{human_cat2_reference} The original human annotation identifying a potential violation of the student's knowledge profile.

5. Violation Classification Codebook
{violation_codebook}

The definitions and criteria for assigning violation categories.

ANNOTATION PROCEDURE

Read the student's system prompt and knowledge profile to identify the assigned knowledge, misconceptions, reasoning procedures, and other relevant cognitive constraints.

Review the preceding conversation history to understand the instructional context.

Read the human reference judgment to identify the potentially relevant knowledge-profile requirement. Treat this reference as secondary evidence, not as a substitute for independent evaluation.

Examine the current target student utterance and determine which, if any, knowledge-profile requirements it violates.

Assign the applicable violation categories using only the supplied codebook definitions. Multiple labels may be assigned when supported.

ANNOTATION CRITERIA

A violation label is supported only when the current student utterance conflicts with an existing requirement in the assigned knowledge profile, interpreted in its conversational context.

Focus exclusively on the current target student utterance. Do not classify errors appearing only in earlier responses.

Evaluate knowledge-profile alignment rather than general correctness, completeness, answer quality, or teacher behavior.

Do not infer knowledge-profile requirements that are not established in the supplied system prompt.

Use only the supplied codebook definitions. Do not introduce additional categories.

OUTPUT

Return the applicable violation category labels using the exact identifiers defined in the supplied codebook.
\end{lstlisting}

\subsection{Violation Recurrence Assessment (RQ2–RQ3)}
\label{app:judge-benchmark}

For the case-linked evaluation, each newly generated target response is assessed for profile conformance (\textit{Yes} = conforms; \textit{No} = violates the profile). The four judges are the same models listed above. The separate original-error field, where present, is distinct from the conformance label. The operative portion of the judge system prompt is reproduced below:

\begin{lstlisting}[style=appendixprompt]
You are an independent evaluator of cognitive fidelity in LLM-based learner simulation. Your task is to assess whether a newly generated student response aligns with the assigned student knowledge profile.

INPUTS

1. Student Knowledge Profile (All protocols)
{student_knowledge_profile}

2. Conversation Context (P2-P4 only, omitted for P1)
{protocol_specific_conversation_context}

P1: No original conversation history is provided.

P2: Original conversation history serialized in the system prompt.

P3: Original conversation history in its native message-role structure.

P4: Dynamically regenerated conversation history.

3. Target Teacher Question (All protocols)
{target_teacher_question}

4. Original Student Response (All protocols)
{original_student_response}

5. Original Human Judgment (All protocols)
The original student response was judged as not aligned with the assigned student knowledge profile.

6. Human Judgment Reason (All protocols)
{human_judgment_reason}

7. Newly Generated Student Response (All protocols)
{generated_student_response}

EVALUATION QUESTION

Does the newly generated student response align with the knowledge profile the student is expected to maintain?

EVALUATION CRITERIA

Assess consistency with the assigned knowledge, misconceptions, and reasoning procedures.

Evaluate the response within the conversational context available under the corresponding protocol. For P1, rely only on the assigned knowledge profile and adapted target question when interpreting the new response.

Judge cognitive-profile alignment rather than general correctness, completeness, or pedagogical quality.

A factually correct response may still violate the profile if it demonstrates knowledge or reasoning beyond the assigned cognitive boundaries.

Treat the original human judgment and its reason as reference information, not as evidence that the new response necessarily violates the profile.

Evaluate the newly generated response independently. Do not assume that the original violation recurs.

OUTPUT

Return exactly two binary judgments:

1. label (Yes/No)

Yes: The newly generated response aligns with the assigned student knowledge profile.

No: The newly generated response violates the assigned student knowledge profile.

2. same_error_as_before (Yes/No)

Yes: The newly generated response reproduces the same cognitive-profile violation identified in the original human judgment.

No: The newly generated response does not reproduce that specific violation.

Return only the two judgments in the specified output format.
\end{lstlisting}

\end{document}